\documentclass[a4paper]{article}

\usepackage{INTERSPEECH2016}

\usepackage{graphicx}
\usepackage{amssymb,amsmath,bm}
\usepackage{textcomp}
\usepackage{enumitem}
\usepackage{arabtex}
\usepackage{url}
\usepackage[table,xcdraw]{xcolor}
\usepackage{placeins}
\usepackage{float}
\usepackage{esvect}
\restylefloat{table}
\usepackage{placeins}
\usepackage[export]{adjustbox}
\usepackage{color,soul}
\usepackage{fixltx2e}
\usepackage{hyperref} 
\hypersetup{
  colorlinks   = true, 
  urlcolor     = black, 
  linkcolor    = black, 
  citecolor   = black 
}
\usepackage{kbordermatrix}
\usepackage{booktabs}

\usepackage{graphics}

\newcommand{\ra}[1]{\renewcommand{\arraystretch}{#1}}

\ninept

\title{Automatic Dialect Detection in Arabic Broadcast Speech}

\makeatletter
\def\name#1{\gdef\@name{#1\\}}
\makeatother \name{\small{{\em Ahmed Ali$^{1,5}$, Najim Dehak$^{2,3}$, Patrick Cardinal$^{4}$, Sameer Khurana$^{1}$, Sree Harsha Yella$^{2}$, James Glass$^2$, Peter Bell$^5$, Steve Renals$^5$}}}
\address{$^1$Qatar Computing Research Institute, HBKU, Doha, Qatar \\
  $^2$ MIT Computer Science and Artificial Intelligence Laboratory (CSAIL), Cambridge, MA, USA\\
  $^3$ JHU Center for Language and Speech Processing (CLSP), Baltimore, MD, USA\\
  $^4$ \'Ecole de technologie sup\'erieure, D\'epartement de G\'enie Logiciel et des TI, Montr\'eal, Canada\\
  $^5$Centre for Speech Technology Research, University of Edinburgh, UK \\
  {\small \tt \{amali,skhurana\}@qf.org.qa, najim@jhu.edu}
}

\begin{document}

  \maketitle
  \begin{abstract}
  In this paper, we investigate different approaches for dialect identification in Arabic broadcast speech. These methods are based on phonetic and lexical features obtained from a speech recognition system, and bottleneck features using the i-vector framework. We studied both generative and discriminative classifiers, and we combined these features using a multi-class Support Vector Machine (SVM). We validated our results on an Arabic/English language identification task, with an accuracy of 100\%. We also evaluated these features in a binary classifier to discriminate between Modern Standard Arabic (MSA) and Dialectal Arabic, with an accuracy of 100\%.  We further reported results using the proposed methods to discriminate between the five most widely used dialects of Arabic: namely Egyptian, Gulf, Levantine, North African, and MSA, with an accuracy of 59.2\%. We discuss dialect identification errors in the context of dialect code-switching between Dialectal Arabic and MSA, and compare the error pattern between manually labeled data, and the output from our classifier. All the data used on our experiments have been released to the public as a language identification corpus. 
  \end{abstract}

  \section{Introduction}
  \label{sec:intro}

The task of Dialect Identification (DID) is a special case of the more general problem of Language Identification (LID). LID refers to the process of automatically identifying the language class for given speech segment or text document.  DID is arguably a more challenging problem than LID, since it consists of identifying the different dialects within the same language class. The importance of addressing DID can be gauged from its growing interest in the Automatic Speech Recognition (ASR) community~\cite{liu2010dialect}. A good DID system can facilitate the identification of dialectal segments from an untranscribed mixed-speech dataset. This process can help reduce the ASR word error rate (WER) for dialectal data by training ASR systems for each dialect, or by adapting the ASR models to a particular dialect.

The natural language processing (NLP) community has aggregated dialectal Arabic into five regional language groups: Egyptian (EGY), North African or Maghrebi (NOR), Gulf or Arabian Peninsula (GLF), Levantine (LAV), and Modern Standard Arabic (MSA). An objective comparison of the varieties of Arabic dialects could potentially lead to the conclusion that Arabic dialects are historically related, but not synchronically, and are mutually unintelligible languages like English and Dutch. Normal vernacular can be difficult to understand across different Arabic dialects~\cite{holes2004modern}. Arabic dialects are thus sufficiently distinctive, and it is reasonable to regard the DID task in Arabic as similar to the LID task in other languages.  Table~\ref{tab:lexical_difference} shows two phrases across the different dialects, it is clear from this example that there are lexical variations across the different dialects which motivates us to consider it.

Two broad LID approaches have been investigated in the literature:  low-level acoustic features, and high-level phonetic and lexical features. In the lexical area, words, roots, morphology, and grammars~\cite{reynolds2008lid,ambikairajah2011language} have been studied. Acoustic features such as shifted delta cepstral coefficients~\cite{dehak2011language} and prosodic features ~\cite{martinez2012ivector} using Gaussian mixture models (GMMs), i-vector representations and support vector machine (SVM) classifiers ~\cite{dehak2011language} have been shown to be effective for LID. More recent work explored the use of frame-by-frame phone posteriors (PLLRs) ~\cite{plchot2014pllr} as new features for LID. New subspace approaches based on non-negative factor analysis (NFA) for GMM weight decomposition and adaptation~\cite{bahari2014non} were also applied to both LID and DID tasks. GMM weight adaptation subspaces seem to provide complementary information to the classical i-vector framework. Finally, phoneme sequence modeling and its n-gram subspace have been studied for both Arabic DID ~\cite{soltau2011modern} and LID ~\cite{soufifar2012discriminative}.

\FloatBarrier
\begin{table}[ht]
\label{lexicalExamples}
\caption{\textit{Lexical examples in Arabic and Buckwalter format.}}
\resizebox{0.5\textwidth}{!}{%
\begin{tabular}{|l|l|l|l|l|l|}
\hline
EGY     & GLF    & LAV           & MSA      & NOR     & Translation \\ \hline
\begin{tabular}[x]{@{}c@{}} \novocalize \RL{AzAyk} \\AzAYk\end{tabular}                                                               &
\begin{tabular}[x]{@{}c@{}} \novocalize \RL{A^slwnk} \\ A\$lwnk\end{tabular}                             & \begin{tabular}[x]{@{}c@{}} \novocalize \RL{A^slwnk / kyfk} \\ kyfk / A\$lwnk \end{tabular}      & \begin{tabular}[x]{@{}c@{}} \novocalize \RL{kyf .hAlk} \\ kyf HAlk\end{tabular}                            &  \begin{tabular}[x]{@{}c@{}} \novocalize \RL{wA^s rAk} \\ wA\$ rAk\end{tabular}                          &
How are you?        \\ \hline

\begin{tabular}[x]{@{}c@{}} \novocalize \RL{Ant fen} \\Ant fyn\end{tabular}       &
\begin{tabular}[x]{@{}c@{}} \novocalize \RL{wynk} \\wynk\end{tabular}                       				  &
\begin{tabular}[x]{@{}c@{}} \novocalize \RL{wynk} \\wynk\end{tabular}                        				  &
\begin{tabular}[x]{@{}c@{}} \novocalize \RL{Ayn Ant} \\ Ayn Ant\end{tabular}           					 &
\begin{tabular}[x]{@{}c@{}} \novocalize \RL{wyn rAk} \\ wyn rAk\end{tabular}             				&
Where are you?        \\ \hline
\end{tabular}
}
\label{tab:lexical_difference}
\end{table}
\FloatBarrier

In this paper we investigate three Vector Subspace Models (VSMs) for Arabic DID based on 1) lexical, 2) phonetic, and 3) i-vectors.  We conduct a thorough feature selection study of these models to better understand their interaction.  A further contribution of this work is the release of an Arabic DID system so others can extend and improve DID performance on this task.\footnote{https://github.com/Qatar-Computing-Research-Institute/dialectID}

  \section{Vector Space Models}
  \label{sec:dialectfeatures}
\subsection{Senone based Utterance VSM}
\label{sec:svsm}
\textbf{Senone} refers to an n-gram phone sequence. In our case $n \le 4$.
VSM construction takes place in two steps:
first, a phoneme recognizer is used to extract the senone~\cite{Hwang} sequence for a given speech utterance. The phoneme sequence is obtained by automatic vowelization of the training text, followed by vowelization to phonetization (V2P). The 36 chosen phonemes cover all the dialectal Arabic sounds. Further details about the speech recognition pipeline, training data, and phoneme set is given in~\cite{ali2014complete}. For the phoneme sequence, we process the phoneme lattice, and obtain the one-best transcription, ignoring silences as well as noisy silences.
Each speech utterance ($\mathbf{u}$) is then represented as a high dimensional sparse vector ($\mathbf{\vv{u}}$):
    \begin{equation}
      \mathbf{\vv{u}} = \left(A(f(u,s_{1})), A(f(u,s_2)), \ldots, A(f(u,s_d))\right), \label{sv}
    \end{equation}
where $f(u,s_{i})$ is the number of times a senone $s_{i}$ occurs in the speech utterance $u$, and  $A$ is the scaling function. We experiment with both an identity scaling function and $tf.idf$ scaling function, commonly used in the field of Natural Language Processing~\cite{Ramos} to downweight the contribution of the words (in our case senones) that occur in almost all documents (in our case utterances), as these words (senones) do not provide any discriminative information about the documents (utterances).  


The vector space is then represented by the matrix, $\mathbf{U_{s}} \in \mathbb{R}^{d \times N}$ (see Fig~\ref{table-semspace}).  This approach and the notation used to define a VSM is directly inspired by the seminal works in the area of VSM of Natural Language in~\cite{Salton1975,Lowe2000,Pado2007} and in LID~\cite{Li2007}.
\begin{figure}[H]
\centering
\small
\[
  \text{$\mathbf{U_{s}}$} = \kbordermatrix{
    & u_1 & u_2 & \dots & u_N \\
    s_1 & A(f(s_1,u_1) & A(f(s_1,u_2) & \ldots & A(f(s_1,u_N) \\
    s_2 & A(f(s_2,u_1) & A(f(s_2,u_2) & \dots & A(f(s_2,u_N) \\
    \vdots & \vdots  & \vdots &\ddots & \vdots \\
    s_d & A(f(s_d,u_1) & A(f(s_d,u_2) & \dots  & A(f(s_d,u_N)
  }
\]
\normalsize
\caption{\textit{Senone-based utterance VSM. Column vectors of the matrix correspond to the speech utterance vector representation formed using equation~\ref{sv}. $d$ is the size of the senone dictionary, and $N$ is the total number of speech utterances in the dialectal speech database.}}
\label{table-semspace}
\end{figure}

\subsection{Word based Utterance VSM}
\label{sec:wordvsm}
The word-based utterance VSM ($\mathbf{U_{w}}$) is constructed in two steps in a manner similar to the senone features:
An ASR system is used to extract the word sequence for each utterance in the speech database. Details about the ASR system can be found in~\cite{ali2014complete}.
Each speech utterance ($\mathbf{u}$) is then represented as a high-dimensional sparse vector ($\mathbf{\vv{u}}$):
    \begin{equation}
      \mathbf{\vv{u}} = \left(A(f(u,w_{1})), A(f(u,w_2)), \ldots, A(f(u,w_{d^{\prime}}))\right) , \label{wv}
    \end{equation}
where $f(u,w_{i})$ is the number of times a word $w_{i}$ occurs in the speech utterance $u$ and $A$ is the scaling function which has the same interpretation as for $\mathbf{U_{s}}$ (above). Vocabulary size was 55k. The tri-gram dictionary size was 580k which we used to construct the word based VSM


\subsection{i-vector-based Utterance VSM}
\label{sec:ivsm}
\subsubsection{Bottleneck Features (BN)}
Recently, bottleneck features extracted from an ASR DNN-based model were applied successfully to language identification~\cite{Ysong2013,Pavel2014,Fred2015}. In this paper, we used a similar bottleneck features configurations as in our previous ASR-DNN system for MSA speech recognition~\cite{Card2015}. This system is based on two successive DNN models. Both DNNs use the same setup of 5 hidden sigmoid layers and 1 linear BN layer, and they were both based on tied-states as target outputs. The senone labels of dimension 3040 are generated by a forced alignment from an HMM-GMM baseline trained on 60 hours of manually transcribed Al-Jazeera MSA news recordings ~\cite{ali2014complete}.
The input to the first DNN consists of 23 critical-band energies that are obtained from Mel filter-bank. Pitch and voicing probability are then added. 11 consecutive frames are then stacked together. The second DNN is used for correcting the posterior outputs of the first DNN. In this architecture, the input features of the second DNN are the outputs of the BN layer from the first DNN. Context expansion is achieved by concatenating frames with time offsets of -10, -5, 0, 5, and 10. Thus, the overall time context seen by the second DNN is 31 frames.

\subsubsection{Modeling}
An effective and well-studied method in language and dialect recognition is the i-vector approach ~\cite{bahari2014non,dehak2011front,dehak2011language}. The i-vector involves  modeling speech using a universal background model (UBM) -- typically a large GMM -- trained on a large amount of data to represent general feature characteristics, which plays a role of a prior on how all dialects look like. The i-vector approach is a powerful technique that summarizes all the updates happening during the adaptation of the UBM mean components to a given utterance. All this information is modeled in a low dimensional subspace referred to as the total variability space. In the i-vector framework, each speech utterance can be represented by a GMM supervector, which is assumed to be generated as follows:
\[M = u + Tv \]
Where $u$ is the channel and dialect independent supervector (which can be taken to be the UBM supervector),
$T$ spans a low-dimensional subspace and $v$ are the factors that best describe the utterance-dependent mean offset. The vector $v$ is treated as a latent variable with the  i-vector being its maximum-a-posteriori (MAP) point estimate. The subspace matrix $T$ is estimated using maximum likelihood on large training dataset. An efficient procedure for training and for MAP adaptation of i-vector can be found in~\cite{kenny2008study}. In this approach, the i-vector is the low-dimensional representation of an audio recording that can be used for classification and estimation purposes. In our experiments, the UBM was a GMM with 2048 components, BN features were used, and the i-vectors were 400-dimensional.

In order to maximize the discrimination between the different dialect classes in the i-vector space, we combine Linear Discriminant Analysis (LDA) and Within Class Co-variance Normalization~\cite{dehak2011language}. This intersession compensation method has been used with both SVM~\cite{dehak2011language} and cosine scoring~\cite{bahari2014non}.

  \section{Dataset}
  \label{sec:data}

 \subsection{Train Data}

The training corpus was collected from the Broadcast News domain in four Arabic dialects (EGY, LAV, GLF, and NOR) as well as MSA. Data recordings were carried out at 16Khz. The recordings were segmented to avoid speaker overlap, removing any non-speech parts such as music and background noise. More details about the training data can be found in~\cite{bahari2014non}. Although the test database came from the same broadcast domain, the recording setup is different. The test data was downloaded directly from the high quality video server for Aljazeera (brightcove) over the period of July 2104 until January 2015, as part of QCRI Advanced Transcription Service (QATS)~\cite{ali2014qcri}.
\begin{table}[pht!]
\centering
\ra{1.3}
\resizebox{48mm}{7mm}{%
\begin{tabular}{@{}llllllll@{}} \toprule Data & EGY & GLF & LAV & NOR & MSA & ENG\\ \midrule Train& 13  & 9.5  & 11 & 9 & 10 & 10 \\
 Test& 2  & 2 & 2 &  2 & 2 & 2\\
\bottomrule \end{tabular}%
}
\caption{\textit{Number of hours of speech available for each dialect.}}\label{tab:data1}
\end{table}
\subsection{Test Data}
\label{testset}
The test set was labeled using the crowdsource platform CrowdFlower, with the criteria to have a minimum of three judges per file and up to nine judges, or 75\% inter-annotator agreement (whichever comes first). More details about the test set and crowdsourcing experiment can be found in~\cite{samn2015crowdsource}. The test set used in this paper differs from that used in~\cite{bahari2014non} for two reasons: First, the crowdsourced data is available to reproduce the results, and thus can be used as a standard test set for Arabic DID; second, the new test set has been collected using different channels, and recording setup compared to the training data, which makes our experiments less sensitive to channel/speaker characteristics.

The train and test data can be found on the QCRI web portal\footnote{http://alt.qcri.org/resources/ArabicDialectIDCorpus/}. Table~\ref{tab:data1} and Table~\ref{tab:data2} present some statistics about the train and the test data.

\begin{table}[pht!]
\centering
\ra{1.3}
\resizebox{48mm}{7mm}{%
\begin{tabular}{@{}llllllll@{}} \toprule Data & EGY & GLF & LAV & NOR & MSA & ENG\\ \midrule Train& 1720  & 1907  & 1059 & 1934 & 1820 & 1649 \\
 Test& 315  & 348  & 238 &  355 & 265 & 452 \\
\bottomrule \end{tabular}%
}
\caption{\textit{Number of speech utterances for each dialect.}}\label{tab:data2}
\end{table}


  \section{Experiments}
  \label{sec:did}
\subsection{Choosing the Best Classifier}

We first studied the best classification approach for the DID task from a set of two generative models: n-gram language model~\cite{Collins} and Naive Bayes~\cite{Ng}, and two discriminative classifiers: linear SVM~\cite{drucker1999support} and Maximum Entropy~\cite{Nigam}.  We measured the performance of each model on the DID task, in the word or lexical-based utterance vector space, which is constructed using the approach mentioned in section~\ref{sec:dialectfeatures}, using identity scaling function $A$, and performing no dimensionality reduction. Hence, the dimensionality of an utterance vector, $\vv{u}$, is the same as the size of the lexicon, which in our case was $55$k. Results can be seen in table~\ref{tab:classifiers}.  As the linear SVM performs the best, it is our choice of classifier for the rest of the experiments.

\begin{table}
\centering
\ra{1.3}
\resizebox{70mm}{13mm}{%
\begin{tabular}{@{}llllll@{}} \toprule Model & ACC & PRC & RCL\\ \midrule
n-gram Language Model& 40.4\%  & 40.2\% & 41.3\% \\
Naive Bayes& 37.9\%  & 37.5\% &  50.2\% \\
Max Ent& 40\%  & 40\% &  40.6\% \\
SVM & {\bf 45.2}\%   & 44.8\%  & 45.4\% \\
\bottomrule \end{tabular}%
}
\caption{\textit{Performance of different classifiers using lexical features, with lexicon size of 55K. ACC, PRC and RCL correspond to accuracy, precision and recall on the test set.}}\label{tab:classifiers}
\end{table}

\subsection{Feature Selection Study}
\label{fab}
 \begin{table*}\centering
 \ra{1.5}
 \resizebox{170mm}{19mm}{%
 \begin{tabular}{@{}rrrrcrrrcrrrcrrr@{}}\toprule
 & \multicolumn{3}{c}{$d = 300$} & \phantom{abc}& \multicolumn{3}{c}{$d = 600$} &
 \phantom{abc} & \multicolumn{3}{c}{$d = 1200$} & \phantom{abc}& \multicolumn{3}{c}{$d = 1600$} \\ \cmidrule{2-4} \cmidrule{6-8} \cmidrule{10-12} \cmidrule{14-16}
 & $ACC$ & $PRC$ & $RCL$ && $ACC$ & $PRC$ & $RCL$ && $ACC$ & $PRC$ & $RCL$  && $ACC$ & $PRC$ & $RCL$\\ \midrule
 $\mathbf{U_{w}^{i}}$  & 38.3 & 41.9 & 39.4 && 41.7 & 44.1 & 42.8 && \textbf{42.9} & 45.6 & 44 && 42.9 & 45 & 43.8\\
 $\mathbf{U_{w}^{tfidf}}$ & 43.3 & 42.7 & 43.5 && 44.6 & 44 & 44.9 && \textbf{45.5} & 45.1 & 45.8 && 21.9 & 20.9 & 21.9\\
 $\mathbf{U_{s}^{i}}$  & 45.2& 44.8& 45.9&& \textbf{45.8} & 45.1 & 46.5 && 45.2 & 44.7 & 45.8\\
 $\mathbf{U_{s}^{tfidf}}$& \textbf{44} & 43.9& 44.7&& 44 & 44.2 & 44.6 && 43.9 & 44 & 44.3\\
 Feature Combination & \textbf{44.8} & 44.2& 45.6&& 44.1& 43.4& 44.8 && 44.8& 44.1& 45.4\\
 \\ \bottomrule
 \end{tabular}%
 }
 \caption{\textit{Accuracy, Precision and Recall for different senone and lexical feature based Vector Spaces. $d$ is the dimensionality of the Vector Space. \textbf{Boldfaced} numbers are the best accuracy for the corresponding vector space, for a corresponding vector space dimensionality $d$. A detailed explanation of feature spaces is given in the feature selection study (section~\ref{fab})}}\label{results-1}
 \end{table*}
 
 \begin{table}[pht!]
 \centering
 \ra{1.5}
 \resizebox{58mm}{10mm}{%
 \begin{tabular}{@{}llllll@{}} \toprule Feature Space & d & ACC & PRC & RCL\\ \midrule $\mathbf{U_{iVec}^{bnf}}$ & 400 &55.3 & 61 & 55.9 \\
  $\mathbf{U_{iVec+LDA+WCNN}^{bnf}}$ &4 & 58.5 & 62.3 & 58.9 \\
  $\mathbf{U_{iVec+LDA+WCNN+LNORM}^{bnf}}$&4 & 58.7 & 61.9 & 59.3 \\
  $\mathbf{U_{iVec+LDA+WCNN}^{bnf} + U_{s}^{i}(600d)}$&604 & 59.2 & 62.7 & 59.5\\
 \bottomrule \end{tabular}%
 }
 \caption{\textit{Accuracy, Precision and Recall for different i-vector based feature spaces. $d$ refers to the dimensionality of the Vector Space. A detailed explanation of feature spaces is given in the feature selection study (section~\ref{fab}).}}\label{tab:results2}
 \end{table}

Here we examine the dialect information captured by the three utterance VSMs explained in section~\ref{sec:dialectfeatures}. We also explore the concatenation of the utterance vector representations, and report the results in Tables~\ref{results-1} and~\ref{tab:results2}. Details about the terms in the results table are given below:
\begin{itemize}
  \itemsep0em
  \item $\mathbf{U_{w}^{i}}$: Refers to the utterance VSM in which each utterance is represented by a vector given by equation~\ref{wv}, where $A$ is chosen to be the identity function. The bases of the vectors are the words in the lexicon. SVD is used to reduce the dimensionality of the utterance Vector Space from $55$k originally, to $300$, $600$, $1200$, $1600$ at which point increase the gain in the classification performance tends to saturate.
  \item $\mathbf{U_{w}^{tf.idf}}$: Same as the previous Utterance VSM, except that $A$ is chosen via $tf.idf$~\cite{Ramos} instead of identity function, which gives us significant improvement in accuracy over the previous vector space.
  \item $\mathbf{U_{s}^{i}}$: Refers to the utterance VSM in which each utterance is represented by a vector given by equation~\ref{sv}, where $A$ is chosen to be the identity function. Utterance vector bases corresponds to senones. Just as with the word-based utterance VSM, we use SVD on the vector space and experiment with different dimensions. The utterance Vector Space constructed using senone features is more discriminative than word-based Vector space.
  \item $\mathbf{U_{s}^{tf.idf}}$: Refers to the same vector space as the previous one, except that $A$ is chosen to be the $tf.idf$ function. $tf.idf$, does not help in the case of senone features.
  \item \textbf{Feature Combination}: Combining the best senone-based utterance VSM, $U_{s}^{i}(600d)$, and the best lexical-based utterance VSM, $U_{w}^{tfidf}(1200d)$, to form a concatenated feature vector representation. SVD is performed to reduce the dimensions of the feature space. Feature combination does not help and hence we conclude that the two vector spaces are capturing similar information.
  \item $\mathbf{U_{iVec}^{bnf}}$: Refers to the utterance VSM, where each utterance is represented by a compact $400d$ i-vector (section~\ref{sec:ivsm}). We use the bottleneck features to train the UBM, which is then used to extract the i-vector. We do not experiment with different i-vector dimensions and take the best dimension reported in~\cite{dehak2011language} for the LID task. The i-vector feature space is significantly more discriminative than previously defined feature spaces.
  \item $\mathbf{U_{iVec+LDA+WCNN}^{bnf}}$: Reducing the dimensionality of the i-vector space using LDA and performing WCNN has been reported to do well in LID tasks~\cite{dehak2011language} and we use the same technique and see a significant improvement in the DID results.
  \item $\mathbf{U_{iVec+LDA+WCNN}^{bnf} + U_{s}^{i}(600d)}$: Finally we concatenate the best senone-based VSM with the best i-vector-based VSM, to form a concatenated vector representation for each utterance and see slight improvements in the results. As the lexical and senone-based representations encode the same information about the dialect, we do not experiment with concatenated lexical and i-vector representations.
\end{itemize}

\subsection{One Vs All classification (Sanity Check)}
We  constructed a senone-based utterance VSM (section~\ref{sec:svsm}) based on 20 hours of speech; 10 hours English (which we got from~\cite{glass2004analysis}) and 10 hours Arabic (randomly sampled from our training data, section~\ref{sec:data}). Binary classification (English vs Arabic) using an SVM classifier, was then performed and it yielded 100\% accuracy on the 1.5 hour test set. The reason to choose the senone-based feature space and not the i-vector-based feature space for classification is to avoid channel mismatch, as the English data came from a different source domain. We did a similar experiment to classify MSA versus all dialectal Arabic and again obtained 100\% classification accuracy.

\subsection{System Output Combination}
We fused the scores of the best senone system and the SVM-based i-vector system. In the fusion steps, the original scores of each system were normalized and combined using the same fusion weights for both systems. This approach yielded a final accuracy of 60.2\%, which is the best performance we achieved. One explanation for this gain is that the error patterns for the two feature spaces are quite different, and we were able to confirm that by analyzing the confusion matrix for each system.

  \section{Discussion}
  \label{cs}

We infer from the confusion matrix in Table~\ref{tab:cm} that GLF and LAV are the most confusable dialect pair. We believe that this is related to the greater lexical similarity between these two dialects (see Table~\ref{tab:lexical_difference}). Note, the confusion matrix is from the best DID system.
We borrowed Table~\ref{tab:expansion2} from previous work~\cite{samn2015crowdsource} on the test set, which shows the amount of time the same speakers switch between dialect and another (mainly MSA, and their own native dialect). For example, in the second row of Table~\ref{tab:expansion2}, there are 200 samples from potential Gulf speakers. After manually labeling, there were 106(53\%) segments labeled as MSA, 82(41\%) validated as GLF, 8(4\%) as LAV, and 4 segments were not identified with enough confidence to be considered. This means more than 50\% of the random GLF speakers data is in-fact MSA speech segments. This is strong evidence for the amount of code-switching between one dialect and MSA from the same speaker.



\begin{table}
\centering
\ra{1.3}
\resizebox{80mm}{16mm}{%
\begin{tabular}{@{}lllllllll@{}} \toprule  & EGY & GLF & LAV & MSA & NOR & Total Truth & PRC\\ \midrule
EGY & 221  & 15 & 57 & 13 & 9 & 315 & 50.3\% \\
GLF & 45 & 121 & 82 & 12 & 5 & 265 & 55.8\% \\
LAV & 74 & 43 & 199 & 18 & 14 & 348 & 46.9\% \\
MSA & 19 & 17 & 20 & 218 & 5 & 279 & 77\% \\
NOR & 80 & 21 & 66 & 22 & 166 & 355 & 83.4\% \\
\#class & 439 & 217 & 424 & 283 & 199 \\
RCL & 70.2\% & 45.7\% & 57.2\% & 78.1\% & 46.8\% \\
\bottomrule \end{tabular}%
}
\caption{\textit{Confusion Matrix for DID.}}\label{tab:cm}
\end{table}


\FloatBarrier
\begin{table}[H]
\centering
\ra{1.3}
\resizebox{50mm}{10.5mm}{%
\begin{tabular}{@{}lllllll@{}} \toprule Expected Dialect & EGY & GLF & LAV & NOR & MSA \\ \midrule
EGY & 65\% & & & & 32\% \\
GLF & & 41\% & 4\% &  & 53\%  \\
LAV & 1\% & 1\% & 53\% &  & 39\%  \\
NOR & 1\% &  &  & 69\% &  28\% \\
\bottomrule \end{tabular}%
}
\caption{\textit{Expected dialect of each speech segment from particular dialectal speakers.}}\label{tab:expansion2}
\end{table}

  \section{Conclusions}
  This paper presents our efforts on automatic dialect identification for Arabic broadcast speech. We have demonstrated a dialect classifier with an accuracy of 60.2\% using system combination. We also achieved 100\% accuracy on two binary classification tasks; MSA vs Dialectal Arabic and English vs Arabic. We studied the potential code-switching pattern in our classifier and its correlation with the manual annotation. Further work for this research is to study the code-switch between MSA and dialectal Arabic without considering speaker diarization or silence between speech segments in what can be called dialect diarization. We shall also study deep neural network approaches of classification to learn a more complex non-linear decision boundary.

  \newpage
  \eightpt
  \bibliographystyle{IEEEtran}

  \bibliography{mybib}


\end{document}